\documentclass{article}

\usepackage[final, nonatbib]{neurips_2022}

\usepackage[utf8]{inputenc} 
\usepackage[T1]{fontenc}    
\usepackage{hyperref}       
\usepackage{url}            
\usepackage{booktabs}       
\usepackage{amsfonts}       
\usepackage{nicefrac}       
\usepackage{microtype}      
\usepackage{xcolor}         
\usepackage{graphicx}
\usepackage{amsmath}
\usepackage{multirow}
\usepackage{caption,subcaption}
\usepackage{pgffor}
\usepackage{multicol}
\usepackage{amsmath}
\newcommand{\inlineeqnum}{\refstepcounter{equation}~~\mbox{(\theequation)}}

\title{KG-NSF: Knowledge Graph Completion with a Negative-Sample-Free Approach}

\author {
	Adil Bahaj\textsuperscript{\rm 1},
	Safae Lhazmir\textsuperscript{\rm 1},
	Mounir Ghogho\textsuperscript{\rm 1}\\
	\textsuperscript{\rm 1} International University of Rabat\\
	adil.bahaj@uir.ac.ma, mounir.ghogho@uir.ac.ma
}

\begin{document}

\maketitle

\begin{abstract}
	
	Knowledge Graph (KG) completion is an important task that greatly benefits knowledge discovery in many fields (e.g. biomedical research). In recent years, learning KG embeddings to perform this task has received considerable attention. Despite the success of KG embedding methods, they predominantly use negative sampling, resulting in increased computational complexity as well as biased predictions due to the closed world assumption. To overcome these limitations, we propose \textbf{KG-NSF}, a negative sampling-free framework for learning KG embeddings based on the cross-correlation matrices of embedding vectors. It is shown that the proposed method achieves comparable link prediction performance to negative sampling-based methods while converging much faster.
\end{abstract}

\section{Introduction}
	In the past few years, large Knowledge Graphs (KGs) have become one of the most valuable resources for improving AI applications, such as information retrieval, question answering, and information extraction. A KG is a multi-relational directed graph that represent real-world knowledge as relational facts (also called triples), which are denoted as $ (h, r, t) $, where $h$ and $t$ correspond to the head and tail entities and $r$ denotes the relation between them, e.g., <covid, causes, headaches>. Freebase, Yago, Gene Ontology, and NELL are examples of large and modern KGs that contain millions of facts. Even though they are successful and popular, their coverage is still far from complete and comprehensive due to the vast amount of facts that could be asserted about our world. However, it is possible to infer new facts from known facts. This motivates the research on knowledge graph completion. The problem of KG completion can be viewed as predicting new triples based on the existing ones. More formally, the goal of KG completion is to predict either the head entity in a given query $ (?, r, t) $ or the tail entity in a given query $ (h, r, ?) $. KG embedding has been proposed to address this issue.
	
	Basically, KG embedding targets to map the entities/relations to dense, low dimensional, and real-valued vectors (called embeddings), capturing the intrinsic structural information of the knowledge graph. Then based on the embeddings, a scoring function $f$ is employed to compute a score to infer whether a candidate triple is a fact. Various scoring functions have been proposed to improve the quality of the embedding. Typical examples include DistMult \cite{yang2015embedding}, SimplE \cite{kazemi2018simple}, ComplEx \cite{trouillon2016complex}, TransE \cite{bordes2013translating}, RESCAL \cite{nickel2011three}, etc. These methods aim to learn a model that distinguishes positive (i.e. true triples) and negative (i.e. false triples) instances based on the optimization of a loss function. The need for negative sampling comes from the fact that there are only positive triples in a KG. Negative sampling under the closed world assumption considers all non-existing triples to be negative by default, which may lead to a significant amount of false negatives. In addition, it has been shown that the hardness of negative examples can lead to the so-called representation collapse \cite{wu2017sampling}. Indeed, extremely hard and extremely easy negative examples both lead to representation collapse, which makes the process of negative sampling a critical part of KG embedding.
	
	In this paper, we propose a new approach for KG completion which does not require negative sampling, which we call \textbf{KG-NSF}. It learns embeddings for relations and entities for the purposes of KG completion without using negative examples and without optimizing objectives that depend on the score function directly.
	
	The contributions of this work can be summarized as follows:
	\begin{itemize}
		\item We develop a negative sample-free approach for learning KG embeddings based on the cross-correlation matrices of the embedding vectors. In addition to not requiring negative sampling, the proposed approach converges faster than existing methods, thus exhibiting reduced computational complexity. Further, our approach achieves link prediction performance comparable to that of negative sampling-based methods.
		\item We show that the loss functions used in traditional KG embedding methods too can be expressed in terms of the cross-correlation matrices of the embedding vectors, but only through their diagonal elements. 
	\end{itemize}
	
\section{Related Literature}
	\subsection{Knowledge Graph Embedding}
\begin{figure}
			\centering
			\includegraphics[scale=0.35, page = 2]{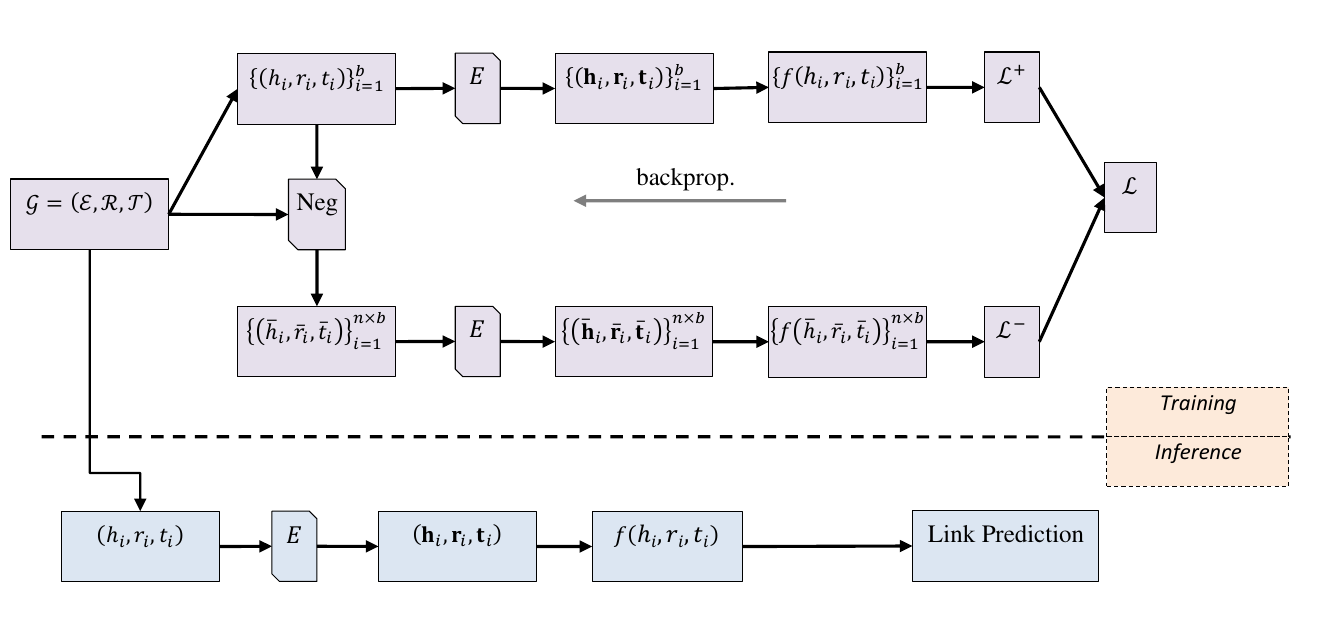}
			\caption{Traditional negative sampling-based KG embedding. $ E $ is the embedding process, which transforms entities and relations to their respective vectorial representations. $ \text{Neg} $ is a negative sampling module that generates $ n $ negative examples for each triple. $ f $ is a scoring function. The bold letters $\mathbf{h}$, $\mathbf{t}$ and $\mathbf{r}$ are embeddings of $ h $, $ t $ and $ r $ respectively.}
			\label{fig:simpKGE}
		\end{figure}

		A KG is a data structure that represents complex semantic relations between conceptual entities. Formally, a KG $\mathcal{G}$ can be presented as the tuple $ (\mathcal{E},\mathcal{R},\mathcal{T}) $, where $\mathcal{E}$ is the set of entities, $\mathcal{R}$ is the set of relations, and $\mathcal{T}=\{(h,r,t);h,t\in\mathcal{E}, r\in \mathcal{R}\}$ is the set of triples. KG embedding aims at learning low dimensional representations of entities and relations in a KG while preserving their semantic properties. These embeddings can be used for various tasks such as link prediction, triple classification, clustering, and entity alignment. In general, KG embedding methods have the following common core elements \cite{wang2017knowledge}: a) a representation space, which is the low-dimensional space in which the entities and relations are represented; most methods use $ \mathbb{R}^{d},d>1 $ as the representation space \cite{bordes2013translating, wang2014knowledge, ji2015knowledge, lin2015learning}, while other methods use different spaces, e.g. the complex space $ \mathbb{C}^{d} $ \cite{trouillon2016complex}, the quaternion space $ \mathbb{H}^{d} $ \cite{zhang2019quaternion}, the box space \cite{vilnis2018probabilistic}; b) a scoring function $ f $, which is a function that assigns a score to each triple in the dataset; this score reflects the goodness of that triple; c) a training process, which is the method that is used to learn the embeddings of relations and entities to have "good" representations given a score function. The training process is characterized by two main features. The first feature is the negativity assumption which designates what triples should be considered false (i.e. negative) and what triples should be considered true (i.e. positive). There are two prevailing assumptions: the closed world assumption, which signifies that every triple that is not explicitly stated in the data is a negative triple, and the open world assumption which states that all triples in a KG are positive and all non-observed triples can either be true or false. The second feature is the loss function (or training objective) $\mathcal{L}$ to be optimized; it is computed using positive and negative examples generated using the considered negativity assumption. In general, the training objective can be expressed as follows:
		\begin{equation}\label{key}
			\mathcal{L}=\sum_{(h,r,t)\in\mathcal{T}}\left(\mathcal{L}^{+}\left(f(h,r,t)\right)+\sum_{(\bar{h},\bar{r},\bar{t})\in \text{Neg}(h,r,t;n)}\mathcal{L}^{-}\left(f(\bar{h},\bar{r},\bar{t})\right)\right)
		\end{equation}
		where $ \text{Neg}(h,r,t;n) $ is a negative sampling function that generates $ n $ negative examples per positive example $ (h,r,t) $, and $\mathcal{L}^{+}$ and $\mathcal{L}^{-}$ are the loss functions for positive and negative triples respectively. Figure \ref{fig:simpKGE} summarizes the traditional KG embedding approach. Our approach differs from these methods in many aspects. First, we avoid using negative sampling which can account for a substantial amount of computational power when training \cite{hajimoradlou2022stay} (e.g. percentage of time spent on negative sampling for each training epoch of SimplE is 69.3\% \cite{hajimoradlou2022stay}). Second, our objective does not include a direct expression of the score function which calculates the likelihood of the existence of a link. Instead, we derive our objective based on the score function in a way that makes the embeddings of heads and entities in their respective triples invariant to transformations based on the embedding of relations.
	\subsection{Self-supervised Learning (SSL)}
		In recent years, SSL methods have achieved state-of-the-art performance on many tasks \cite{wu2021self, jaiswal2020survey}. Depending on how negative examples are used, we can categorize these methods into two categories: a) contrastive SSL methods and b) non-contrastive SSL methods. The former is the more traditional approach to SSL, which uses negative examples to make the embeddings of negative examples and positive examples as distant as possible from each other while making the embeddings of positive examples as close as possible to each other. The use of negative examples is motivated by the representation collapse problem, which occurs when models learn trivial representations that optimized the objective function but do not encode any useful information for downstream tasks. Non-contrastive SSL methods \cite{grill2020bootstrap, zbontar2021barlow, bielak2021graph} do not make use of negative sampling, and employ other approaches to avoid representation collapse. Some of these methods use regularization (BYOL \cite{grill2020bootstrap}). Other methods use specially designed objectives such as Barlow Twins (BT) \cite{zbontar2021barlow}. We make use of this loss in our work. \\
		Our training objective is based on the BT loss function, which was introduced in \cite{zbontar2021barlow} as a negative-sampling free pre-training objective for image pretext tasks. It was later applied in a similar manner for graph representation learning \cite{bielak2021graph} and videos \cite{wang2022disentangled}. To the best of our knowledge, this loss function has not been used before in the context of KG completion.
		
		There are fundamental differences in the way we use the BT loss in our objective function. First, we do not use our objective function to train representations on a pretext text, which would later be used in a downstream task. Our pretext task and downstream task are the same, i.e. link prediction. The models trained using our objective function are ready for inference immediately after training. Second, we do not apply any kind of augmentations on the KG. Consequently, our representations are not invariant to the augmentation of the data. Third, our objective function is a combination of two BT losses instead of one.
	\subsection{Barlow Twins and Variants}
		The BT loss was introduced in \cite{zbontar2021barlow} for non-contrastive self-supervised learning of image representations. The objective of this loss is to learn invariances by discriminating instances and features rather than positive instances and negative instances. This was achieved through the calculation of the cross correlation matrix $\mathcal{C}$ between two tensors $ \mathbf{X}=[x_{ij}]_{1\le i\le b, 1\le j\le d} $ and $ \mathbf{Y}=[y_{ij}]_{1\le i\le b, 1\le j\le d} $ ($ b $ is the batch size and $ d $ is the dimension of the embedding), whose elements can be expressed as follows:
		\begin{equation}\label{key}
		\mathcal{C}_{ij}=\frac{\sum_{b}x_{b,i}y_{b,j}}{\sqrt{\sum_{b}(x_{b,i})^{2}}\sqrt{\sum_{b}(y_{b,j})^{2}}}
		\end{equation}
		The BT loss consists of the combination of an invariance component and a redundancy component: 
		\begin{equation}
		\label{eq:barlowtwins}
		\mathcal{L}_{\text{BT}}(\mathbf{X},\mathbf{Y})=\sum_{i}(1-\mathcal{C}_{ii})^{2}+\lambda\sum_{i}\sum_{j\neq i}\mathcal{C}_{ij}^{2}
		\end{equation}
		Minimizing the BT loss function $ \mathcal{L}_{BT} $ (Equation \ref{eq:barlowtwins}) forces the cross-correlation matrix to be as close as possible to the identity matrix. Minimizing the invariance component pushes the embeddings to be as invariant as possible to data augmentations by forcing the on-diagonal elements, $ \mathcal{C}_{ii} $, be as close as possible to $ 1 $. Minimizing the redundancy component ensures that the embedding features are as decorrelated (non-redundant) as possible by forcing the off-diagonal elements, $ \mathcal{C}_{ij} $, to be as close as possible to $ 0 $.
		Parameter $ \lambda >0 $ in Eq. \ref{eq:barlowtwins} defines the trade-off between the invariance and redundancy reduction terms when optimizing the BT loss. The authors of \cite{zbontar2021barlow} proposed to use $ \lambda=1/d $, which we adopt in our experiments.\\
	
		The BT loss has its theoretical grounding in Information Bottleneck Theory \cite{tishby2000information}, which supposes that the embedding vectors are drawn from a Gaussian distribution. This assumption is too strong for most practical cases and may not be achievable in some cases. \cite{tsai2021note} remedied this problem by interpreting the BT loss as an instance of negative-sample-free contrastive learning, which connected the BT loss to the Hilbert-Schmidt Independence Criterion (HSIC) \cite{gretton2012kernel}. Representations are learned by maximizing HSIC for augmented views of data points. This can be achieved by minimizing the following loss function:
		\begin{equation}
			\label{eq:hsic}
		\mathcal{L}_{HSIC}=\sum_{i}(1-\mathcal{C}_{ii})^{2}+\lambda\sum_{i}\sum_{j\neq i}(1 +\mathcal{C}_{ij})^{2}
		\end{equation}
		We utilize both of these losses in our experiments. \\
		In addition, we utilize the batch normalization method ShuffledDBN (Shuffled Decorrelated Batch Normalization) described in \cite{hua2021feature}, which was found to increase feature decorrelation and improve the performance of Barlow twins-based methods.
	\subsection{KG Embedding without Negative Sampling}
\begin{figure}[ht!]
			\centering
			\includegraphics[scale=0.35, page = 4]{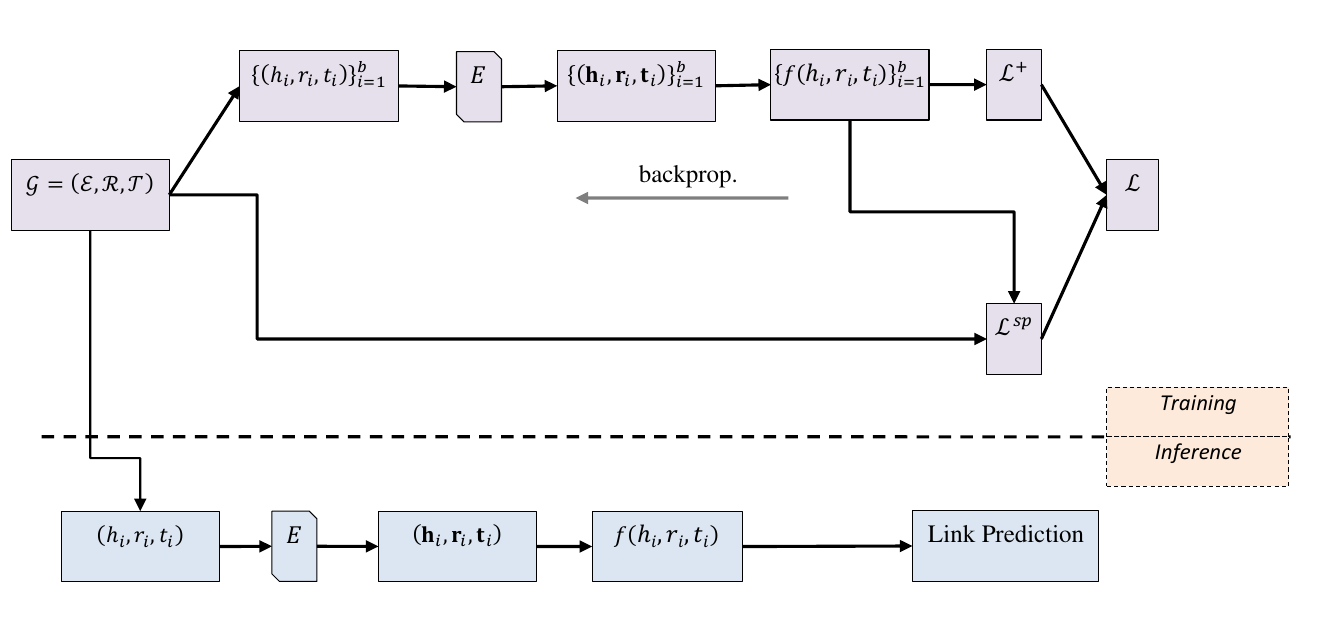}
			\caption{\textit{Stay Positive} Approach.}
			\label{fig:staypositive}
		\end{figure}

		To our knowledge, there is only one method that explores KG embedding without negative sampling \cite{hajimoradlou2022stay}. The objective function in \cite{hajimoradlou2022stay} is defined as follows:
		\begin{equation}\label{key}
			\mathcal{L}=\sum_{(h,r,t)\in\mathcal{T}}\mathcal{L}^{+}(f(h,r,t))+\lambda\mathcal{L}^{sp}(f)
		\end{equation}
		where the first term of this objective function penalizes low scores for the positive examples, while $\mathcal{L}^{sp}$ ($ sp $ stands for \textit{stay positive}) tries to prevent the model from generating high scores for all the triples; $\lambda$ is a hyperparameter. In \cite{hajimoradlou2022stay},  $\mathcal{L}^{sp}$ is defined as:
		\begin{equation}\label{key}
			\mathcal{L}^{sp}(f)=\left\|\sum_{h\in \mathcal{E}}\sum_{r\in \mathcal{R}}\sum_{t\in \mathcal{E}}f(h,r,t)-\psi|\mathcal{E}|^{2}|\mathcal{R}|\right\|
		\end{equation}
		where $\psi$ is a hyperparameter. Figure \ref{fig:staypositive} summarizes the approach in \cite{hajimoradlou2022stay}. Our approach differs from this work in many aspects. First, our objective does not use an explicit expression of the score function, which makes the representations more robust to biases that the score function might induce. Second, we do not solve the representation collapse problem using regularization. We use instead an objective function based on the BT loss. Third, even though the approach in \cite{hajimoradlou2022stay} does not explicitly use negative sampling, the regularization term is computed over all the possible triples in the KG (i.e. positive and negative triples), which is avoided in our approach since our objective function is computed using a batch of positive triples only.
\section{Methodology}
	\label{sec:methodology}
	Motivated by the self-supervised nature of KG embedding, we propose a new framework for KG embedding without the use of negative examples. The overall pipeline of our framework is shown in Figure \ref{fig:framework}. The processing steps can be described as follows:

	\begin{figure}[ht]
		\centering
		\includegraphics[scale=0.35,page=3]{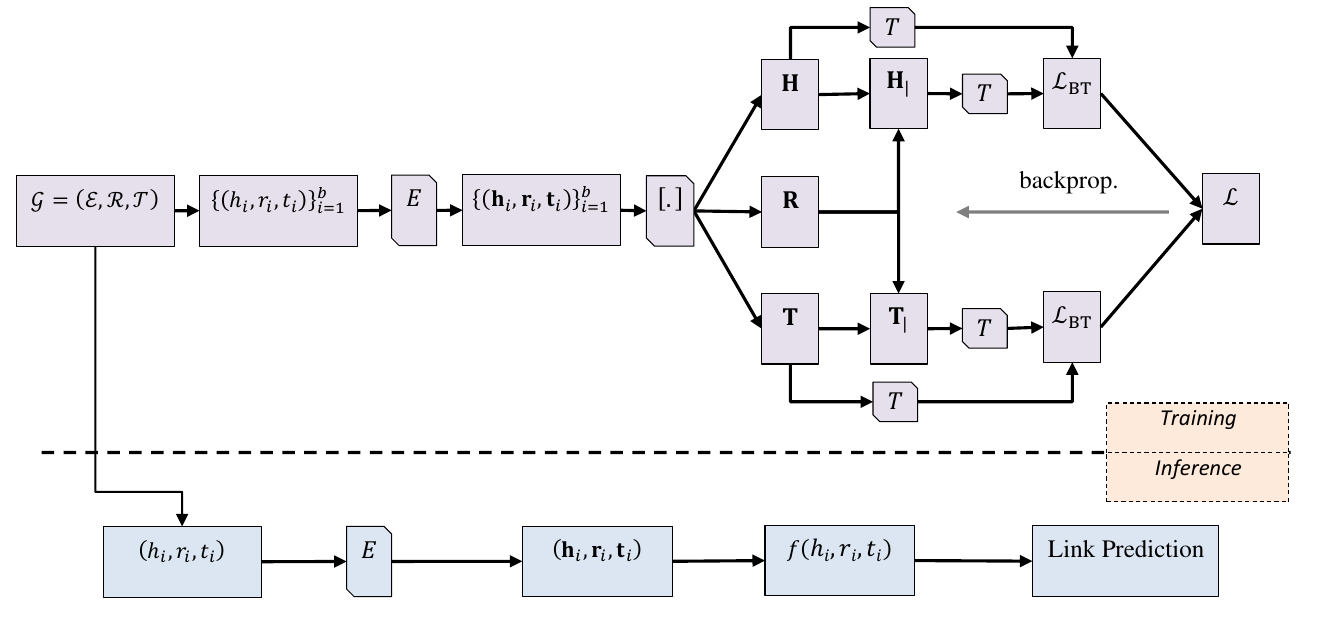}
		\caption{KG-NSF processing pipeline. $ [\cdot] $ is the concatenation operator. $ T $ is a process that transforms the embedded vectors from their original space to a new space.}
		\label{fig:framework}
	\end{figure}
	\paragraph{Knowledge Graph ($ \mathcal{G} $)}
		We present a knowledge graph $\mathcal{G}$ as a triple $ (\mathcal{E},\mathcal{R},\mathcal{T}) $, where $\mathcal{E}$ is the set of entities, $\mathcal{R}$ is the set of relations and $\mathcal{T}=\left\{(h,r,t)| h,t \in\mathcal{E}, r\in\mathcal{R}\right\}$ is the set of triples ($ h \equiv head$, $ t \equiv tail$ and $ r \equiv relation$ ). Each entity/relation $ u $ has an embedding $\mathbf{u}$, which is obtained from a standard embedding layer.
	\paragraph{Embedding ($ E $)}
		An embedding layer is a mapping that transforms the entities/relations from their symbolic form to a vector space. In this work, we represent the embedding vectors of entities and relations with bold letters. Specifically, we represent an embedding of a head $ h $ and tail $ t $ entities with $\mathbf{h}$ and $\mathbf{t}$ respectively. The embedding for the relation $ r $ is presented as $\mathbf{r}$.
	\paragraph{Transformation ($ T $)}
		The transformation module is used to transform the different matrices from their original embedding space to another space with some desired properties. In our work, we experimented with the linear transformation and ShuffledDBN \cite{hua2021feature}.
	\paragraph{Score Function ($ f $)}
		In traditional KG embedding, a score function is used in order to quantify the likelihood of the existence of a triple. The literature contains a multitude of score functions. In this work we focused on the following score functions:
		\begin{itemize}
			\item \textbf{TransE}: TransE is one of the earliest KG embedding methods. It uses a translation based approaches. Its score function is: $f(h,r,t)=-\|\mathbf{h}+\mathbf{r}-\mathbf{t}\|_{p} \inlineeqnum\label{eq:transeSF} $, where $ p\in\{1,2\} $,$ \mathbf{h} $, $ \mathbf{r} $ and $\mathbf{t}$ are the embeddings of $ h $, $ r $ and $ t $ respectively. $ \|\cdot\|_{p} $ is the $ L_{p} $ norm.
			\item \textbf{DistMult}: DistMult also belongs to the family of bi-linear models. It has the following score function: $ f(h,r,t)=\sum_{i}\left[\mathbf{h}\odot \mathbf{r} \odot \mathbf{t}\right]_{i} \inlineeqnum\label{eq:distmultSF}$,	
			where $ \odot $ is the element-wise product, and the $ i $ index is along the feature dimension of the vector resulting from the element-wise product of the different features.
		\end{itemize}
	\paragraph{Loss ($\mathcal{L}$)}
		The general expression of the proposed loss function is:
		\begin{equation}
			\label{eq:tbt}
			\mathcal{L}(\{(h,r,t)\}_{i=1}^{b})=\mathcal{L}_{BT}(\mathbf{H}_{|},\mathbf{T})+\mathcal{L}_{BT}(\mathbf{H},\mathbf{T}_{|}),
		\end{equation}
		where $ \mathcal{L}_{BT} $ is the BT loss, $ b $ is the batch size,  $\mathbf{H}=\left[\mathbf{h}_{i}\right]_{i}\in \mathbb{R}^{b\times d_{e}}$ is a matrix were each row represents the embedding of a head entity in the batch, $\mathbf{T}=\left[\mathbf{t}_{i}\right]_{i}\in \mathbb{R}^{b\times d_{e}}$ is a matrix were each row represents the embedding of a tail entity in the batch, and $\mathbf{R}=\left[\mathbf{r}_{i}\right]_{i}\in \mathbb{R}^{b\times d_{r}}$ is a matrix were each row represents the embedding of a relation in the batch, $ \mathbf{H}_{|} $ is a function of $ \mathbf{H}$ and $ \mathbf{R}$, and $ \mathbf{T}_{|} $ is a function of $ \mathbf{T}$ and $ \mathbf{R}$; these functions are designed according to the score function used to evaluate the triples during inference, as described next. \\

		The score function can be re-written in two forms: $ f(h,r,t)=\begin{cases}
        	p_{1}(g_{1}(h,r),g_{1}^{'}(t))\\
        	p_{2}(g_{2}^{'}(h),g_{2}(t,r))
    	\end{cases} $, where
		$ p_{1} $, $ p_{2} $, $ g_{1} $, $ g_{2} $, $ g_{1}' $ and $ g_{2}' $ have expressions that depend on the kind of score function. For examples, in the case of TransE, $ p_{1}(x,y)=p_{2}(x,y)=||x-y||_{p} $, $ g_{1}(h,r)=\mathbf{h}+\mathbf{r} $, $ g_{1}'(t)=\mathbf{t} $, $ g_{2}(t,r)=\mathbf{t}-\mathbf{r} $, and $ g_{2}'(h)=\mathbf{h} $. In the case of TransH, $ p_{1}(x,y)=p_{2}(x,y)=||x-y||_{p} $, $ g_{1}(h,r)=(\mathbf{h}-\mathbf{w}_{r}^{\top}\mathbf{h}\mathbf{w}_{r})+\mathbf{r} $, $ g_{1}'(t)=\mathbf{t}-\mathbf{w}_{r}^{\top}\mathbf{t}\mathbf{w}_{r} $, $ g_{2}(t,r)=(\mathbf{t}-\mathbf{w}_{r}^{\top}\mathbf{t}\mathbf{w}_{r})-\mathbf{r} $, and $ g_{2}'(h)=\mathbf{h}-\mathbf{w}_{r}^{\top}\mathbf{h}\mathbf{w}_{r} $. In the case of DistMult we have $ p_{1}(x,y)=p_{2}(x,y)=x\odot y $, $ g_{1}(h,r)=\mathbf{h}\odot\mathbf{r} $, $ g_{1}'(t)=\mathbf{t} $, $ g_{2}(t,r)=\mathbf{t}\odot\mathbf{r} $, and $ g_{2}'(h)=\mathbf{h} $. Given these two forms of the score function, we define $\mathbf{H}_{|}$ and $\mathbf{T}_{|}$ as follows: $\mathbf{H}_{|}=\left[g_{1}(h_{i},r_{i})\right]_{i=1}^{b}\in \mathbb{R}^{b\times d_{e}}$ and $\mathbf{T}_{|}=\left[g_{2}(t_{i},r_{i})\right]_{i=1}^{b} \in \mathbb{R}^{b\times d_{e}}$. Hence, considering this reformulation, the only difference in training our models resides in the choice of $g_1$ and $g_2$ through $\mathbf{H}_{|}$ and $\mathbf{T}_{|}$. For the score functions considered in this paper, we choose the following expressions: 
		\begin{itemize}
           \item 
		\textbf{NSF-TransE}: ~~  $ \begin{matrix} \mathbf{H}_{|}=\mathbf{H} + \mathbf{R},&\mathbf{T}_{|}=\mathbf{T} - \mathbf{R} \end{matrix}  ~~  ~~  \inlineeqnum\label{eq:btTransELoss}$
		\item \textbf{NSF-DistMult}: ~~ $ \begin{matrix} \mathbf{H}_{|}=\mathbf{H}\odot\mathbf{R},&\mathbf{T}_{|}=\mathbf{R}\odot\mathbf{T} \end{matrix}   ~~  ~~  \inlineeqnum\label{eq:btdistmultLoss}$.
		\end{itemize}
		
	It is worth pointing out that this approach can be applied directly to other score functions. For example, all translation based models (e.g. TransH \cite{wang2014knowledge}, TransR \cite{lin2015learning}, TransD \cite{ji2015knowledge} etc) can be implemented in a manner analogous to that of TransE, since the only difference is in the expression of the above-mentioned functions. Similarly, any bilinear model (RESCAL \cite{nickel2011three}, SimplE \cite{kazemi2018simple} etc) can be implemented in a way similar to that of NSF-DistMult.
	The design of the proposed loss is motivated by the self-supervised nature of the link prediction task. In fact, link prediction does not require any kind of direct supervision, but rather the positive and negative links are extracted from the data itself (i.e. links that exist are positive, links that do not may be considered negative). Consequently, the kinds of invariances that are learned by the embeddings are intrinsic to the data. For example, TransE tries to minimize a distance ($L1$ or $L2$) betwen $h+r$ and $t$, which can be interpreted as learning entity embeddings invariant to relation translation, which is also what our loss tries to do using the BT loss. Analogously to the invariance term in BT loss, which learns embeddings invariant to data augmentation, in our loss it learns embeddings invariant to relation translation. In methods such as TransE, negative sampling is used to avoid the representation collapse problem \cite{wu2017sampling}. In the proposed method, the second term of BT loss, i.e. the redundancy term, avoids this problem as it forces the features to be as uncorrelated as possible, thus promoting diversity of the embedding representations. Consequently, our approach replaces negative examples with feature decorrelation to avoid the collapsing problem.

	In the appendix. we show that the original TransE and DistMult loss functions too can be expressed as functions of the (non-standardized) cross-correlation matrices of embedding vectors for positive and negative triples, but only thorough their diagonal elements. The proposed BT loss-based method uses both {\em diagonal and off-diagonal} terms of the cross-correlation matrices matrices for positive triples only to learn good (non-trivial) embeddings.
\section{Experiments}
	\label{sec:experiments}
	In our main line of experiments we are interested in quantifying the efficiency of the models proposed in the previous section for the task of link prediction. Other experiments are conducted to highlight other interesting characteristics of our method. Our implementation is based on TorchKGE \cite{arm2020torchkge} python package. All the models were trained on a Geforce RTX 3090 GPU with 24G of memory.
	\subsection{Datasets}
		\label{sec:datasets}
		We choose established datasets in KG completion literature so that we can compare them to our methods. Table \ref{tab:datastats} shows some statistics of the datasets. We used the following datasets: a) \textit{FB15K} which is a subset of a dataset called Freebase \cite{bollacker2008freebase}. Freebase is a large knowledge base that consists of large amounts of facts. We used the split that exit bt default in TorchKGE. b) \textit{WN18} a subset of a dataset called WordNet \cite{miller1995wordnet}. It is a large lexical knowledge graph. Entities in WordNet are synonyms that express distinct concepts. Relations in WordNet are conceptual-semantic and lexical relations. We used the split that exit bt default in TorchKGE. c) \textit{WN18AM} and d) \textit{FB15k-237AM} are subsets of the WN18RR \cite{dettmers2018convolutional} and FB15k-237 \cite{toutanova2015observed} datasets respectively. They were introduced in \cite{hajimoradlou2022stay} after the authors noticed that WN18RR and FB15k-237 training data was missing some entities that exist their respective validation and test splits. We used the same splits introduced by the authors.
		\begin{table}
			\centering
			\scalebox{0.6}{
			\begin{tabular}{|c|c|c|c|c|}
				\hline
				&FB15K&WN18&FB15k-237AM&WN18AM\\
				\hline
				\#Entities&14951&40943&14505&40559\\
				\hline
				\#Relations&1345&18&237&11\\
				\hline
				\#Train Triples&483142&141442&272115&86835\\
				\hline
				\#Validation Triples&50000&5000&17526&2824\\
				\hline
				\#Test Triples&59071&5000&20438&2924\\
				\hline
			\end{tabular}}
			\caption{Dataset statistics.}
			\label{tab:datastats}
		\end{table}
	\subsection{Evaluation Protocol}
		\subsubsection{Experimental Setup}
			\label{sec:expsetup}
			We use the same embedding generating model architecture for all methods. The model's parameters are optimized using negative sampling-based methods and our approach. We experiment with various batch sizes, embedding dimensions and learning rates, following the same line of experimental setups as the methods to which we compare our approach. In our experiments we select the learning rate (lr) among $ \{0.001,0.0005,0.0001\} $, the batch size $ b $ among $ \{100,200,500,1000,2000,4000\} $, and the embedding dimension ($ d=d_{e}=d_{r} $) among $ \{50,100,150,200,300,400,500\} $. We used the Adam optimizer \cite{kingma2014adam}. The optimal parameters are determined by the filtered $ MRR $ metric calculated on the test set. Tables \ref{tab:mainExperiments} and \ref{tab:mainExperiments2} show and compare the performances of the models trained with negative sampling and our models on multiple datasets. In the experiments where we use ShuffledDBN, we fixed its group size to be 5 (see \cite{hua2021feature} for more details on ShuffledDBN).
		\subsubsection{Evaluation Metrics}
			The learned $ f(h,r,t) $ is evaluated in the context of link prediction. As in other works \cite{bordes2013translating, wang2014knowledge, ji2015knowledge, lin2015learning} , for each triple $ (h,r,t) $, we first take $ (?,r,t) $ as the query and obtain the filtered rank on the head: $ 	\text{rank}_{h}=\left|\left\{e\in \mathcal{E}:(f(e,r,t)\ge f(h,r,t))\wedge((e,r,t)\notin \mathcal{S}_{tra}\cup \mathcal{S}_{val}\cup \mathcal{S}_{tst})\right\}\right|+ 1 $, where $ \mathcal{S}_{tra} , \mathcal{S}_{val}, \mathcal{S}_{tst}$ are the training, validation, and test triples,respectively. Next we take $ (h,r,?) $ as the query and obtain the filtered rank on the tail: $ \text{rank}_{t}=\left|\left\{e\in\mathcal{E}:(f(h,r,e)\ge f(h,r,t))\wedge((h,r,e)\notin \mathcal{S}_{tra}\cup \mathcal{S}_{val}\cup \mathcal{S}_{tst})\right\}\right|+1 $. The following metrics are computed from both the head and tail ranks on all triples: (i) Mean reciprocal ranking (MRR):	$ MRR=\frac{1}{2|\mathcal{S}|}\sum_{(h,r,t)\in\mathcal{S}}\left(\frac{1}{\text{rank}_{h}}+\frac{1}{\text{rank}_{t}}\right) $. (ii) $ H@k $: ratio of ranks no larger than $ k $, i.e., $ H@k=\frac{1}{2|\mathcal{S}|}\sum_{(h,r,t)\in\mathcal{S}}\left(\mathbb{I}(\text{rank}_{h}\le k) +\mathbb{I}(\text{rank}_{t}\le k)\right) $, where $ \mathbb{I}(a) = 1 $ if $ a $ is true, otherwise $ 0 $. (iii) $ MR $: mean rank: $ MR= \frac{1}{2|\mathcal{S}|}\sum_{(h,r,t)\in\mathcal{S}}\left(\text{rank}_{h}+\text{rank}_{t}\right) $. The larger the $ MRR $ or $ H@k $, the better is the embedding. The smaller the $ MR $, the better the embedding.
	\subsection{Results}
		
		NSF-TransE (w/o SDBN) and NSF-DistMult (w/o SDBN) are the models trained using our approach without the ShuffledDBN layer. In NSF-TransE (w/ SDBN) and NSF-DistMult (w/ SDBN) we transform $ \mathbf{H}_{|} $, $\mathbf{H}$, $\mathbf{T}$, and $\mathbf{T}_{|}$ using the ShuffledDBN approach described in \cite{hua2021feature} before feeding them to the BT loss. We compare our models to TransE \cite{bordes2013translating} and DistMult \cite{yang2015embedding} trained using negative sampling on the FB15k and WN18 datasets. The results reported for these two methods are taken the corresponding  original papers. The performance of the baselines and our models on the FB15k and WN18 datasets is shown in table \ref{tab:mainExperiments}. In addition, we conducted a set of experiments to compare the performance of our approach with that of the only negative-sample-free approach in the literature reported in \cite{hajimoradlou2022stay}; The results are shown in Table \ref{tab:mainExperiments2}.
		
		We observe that the performance of NSF-DistMult and NSF-TransE variants surpass that of DistMult and TransE on the FK15k and FK15k-237AM datasets, while having a comparable performance on the WN18 and WN18AM datasets. In addition, compared to SP-DistMult, NSF-DistMult has a better performance on the FK15k-237AM dataset and a comparable performance on the WN18AM dataset (i.e. higher $ MRR $ and lower $ H@1 $). Compared to SimplE, NSF-DistMult and NSF-TransE variants have better performance on the FK15k-237AM and a comparable performance on the WN18AM dataset (i.e. higher $ H@1 $ and lower $ MRR $). Compared to SP-SimplE, NSF-TransE (W/ SDBN) have a better $ MRR $ a lower $ H@1 $. The bad performance of NSF-TransE on the WN18 and WN18AM datasets can be attributed to the nature of relationships that the triples in these datasets have. In fact, since (in both datasets) the number of relations is very low, while the number of triples is very high, there is a high incidence of N-1, 1-N and N-N relations, which are known to be one of TransE's weaknesses \cite{wang2014knowledge}. All the models have converged significantly faster than the ones trained using negative sampling (see supplementary material).
		
		\begin{table}[ht]
			\centering
			\scalebox{0.6}{
			\begin{tabular}{|c|c|c|c|c|c|c|c|c|c|c|c|c|c|c|c|c|}
				\hline
				&\multicolumn{8}{c}{FB15K}&\multicolumn{8}{|c|}{WN18}\\
				\cline{2-17}
				&\multicolumn{2}{c}{$ MR $}&\multicolumn{2}{|c}{$ MRR $}&\multicolumn{2}{|c}{$ H@1 $}&\multicolumn{2}{|c}{$ H@10 $}&\multicolumn{2}{|c}{$ MR $}&\multicolumn{2}{|c}{$ MRR $}&\multicolumn{2}{|c}{$ H@1 $}&\multicolumn{2}{|c|}{$ H@10 $}\\
				\cline{2-17}
				&raw& filt.&raw& filt.&raw& filt.&raw& filt.&raw& filt.&raw& filt.&raw& filt.&raw& filt.\\
				\hline
				TransE \cite{bordes2013translating}&\textbf{243}&\textbf{125}&\_&\_&\_&\_&34.9&47.1&\textbf{263}&\textbf{251}&\_&\_&\_&\_&75.4&89.2\\
				\hline
				DistMult \cite{yang2015embedding}&\_&\_&\_&0.35&\_&\_&\_&57.7&\_&\_&\_&\textbf{0.83}&\_&\_&\_&\textbf{94.2}\\
				\hline
				\hline
				NSF-TransE (w/o SDBN)&351&	256&	0.2117&	0.3366&	12.12&	23.70&	39.82&	50.89&	473&	460&	0.2388&	0.3149&	04.32&	07.21&	68.08&	78.59\\
				\hline
				NSF-TransE (w/ SDBN)&253&158&	0.2222&	0.3609&	12.34&	24.45&	43.04&	56.45&	373&	361&	0.2326&	0.3084&	10.59&	16.99&	53.01&	60.53\\

				\hline
				NSF-DistMult (w/o SDBN)& 409& 	311&	0.1930&	0.3265&	10.99&	23.18&	36.75&	49.42&	641&	626&	0.4944&	0.7955&	33.64&	69.05&	79.40&	93.29\\
				\hline
				NSF-DistMult (w/ SDBN)&331&	233&	0.2317&	\textbf{0.4045}&13.24&	29.90&	\textbf{44.41}&		\textbf{59.14}&799&	785&	0.4969&	0.7998&	33.81&	69.75&	\textbf{79.50}&	93.29\\
				\hline
			\end{tabular}}
			\caption{Results on FB15K and WN18.}
			\label{tab:mainExperiments}
		\end{table}
		
		\begin{table}[ht]
			\centering
			\scalebox{0.6}{
				\begin{tabular}{|c|c|c|c|c|c|c|c|c|c|c|c|c|c|c|c|c|}
					\hline
					&\multicolumn{8}{c}{FB15k-237AM}&\multicolumn{8}{|c|}{WN18AM}\\
					\cline{2-17}
					&\multicolumn{2}{c}{$ MR $}&\multicolumn{2}{|c}{$ MRR $}&\multicolumn{2}{|c}{$ H@1 $}&\multicolumn{2}{|c}{$ H@10 $}&\multicolumn{2}{|c}{$ MR $}&\multicolumn{2}{|c}{$ MRR $}&\multicolumn{2}{|c}{$ H@1 $}&\multicolumn{2}{|c|}{$ H@10 $}\\
					\cline{2-17}
					&raw& filt.&raw& filt.&raw& filt.&raw& filt.&raw& filt.&raw& filt.&raw& filt.&raw& filt.\\
					\hline
					DistMult\cite{hajimoradlou2022stay}&	\_&	\_&	\_&	0.256&	\_&	16.0&	\_&	\_&	\_&	\_&	\_&	0.440&	\_&	41.0&	\_&	\_\\
					\hline
					SimplE\cite{hajimoradlou2022stay}&	\_&	\_&	\_&	0.256&	\_&	16.3&	\_&	\_&	\_&	\_&	\_&	0.461&	\_&	25.6&	\_&	\_\\
					\hline
					SP-DistMult\cite{hajimoradlou2022stay}&	\_&	\_&	\_&	0.249&	\_&	18.1&	\_&	\_&	\_&	\_&	\_&	0.448&	\_&	42.3&	\_&	\_\\
					\hline
					SP-SimplE\cite{hajimoradlou2022stay}&	\_&	\_&	\_&	0.250&	\_&	\textbf{18.4}&	\_&	\_&	\_&	\_&	\_&	\textbf{0.470}&	\_&	\textbf{44.5}&	\_&	\_\\
					\hline
					\hline
					NSF-DistMult (w/o SDBN)&	625&	488&	0.1526&	0.2355&	9.14& 		16.17&	27.54&	38.10&	4779&	4765&	0.2832&	0.4526&	16.55& 		40.16&	50.08&	54.77\\
					\hline
					NSF- DistMult (w/ SDBN)&381&249&	0.1785&	0.2614&	11.11&	17.78&	31.78&	42.73&	5263&	5250&	0.2816& 		0.4540&	16.29&	40.37&	50.05&	54.63\\
					\hline
					NSF- TransE (w/o SDBN)&494&	364&	0.1669&	0.2472&	10.36&	17.13&	29.77&	39.63&	1866 &	1854&	0.1459&	0.1808&	2.49&	3.72&	41.09&	45.29\\
					\hline
					NSF- TransE (w/ SDBN)&	386&	254&	0.1766&	\textbf{0.2697}&	10.75&		\textbf{18.36}&	32.04&	43.72&	2046&	2034&	0.1411&	0.1750&	02.49&		03.74&	40.61&	44.71\\
					\hline
			\end{tabular}}
			\caption{Results on FB15k-237AM and WN18AM. SP refers to \textit{stay positive} from \cite{hajimoradlou2022stay}.}
			\label{tab:mainExperiments2}
		\end{table}
\section{Ablation Study}
	
		
		\subsection{Loss Weighting}
\begin{figure}[ht]
				\centering
				\begin{subfigure}[b]{0.4\textwidth}
					\includegraphics[width=\textwidth]{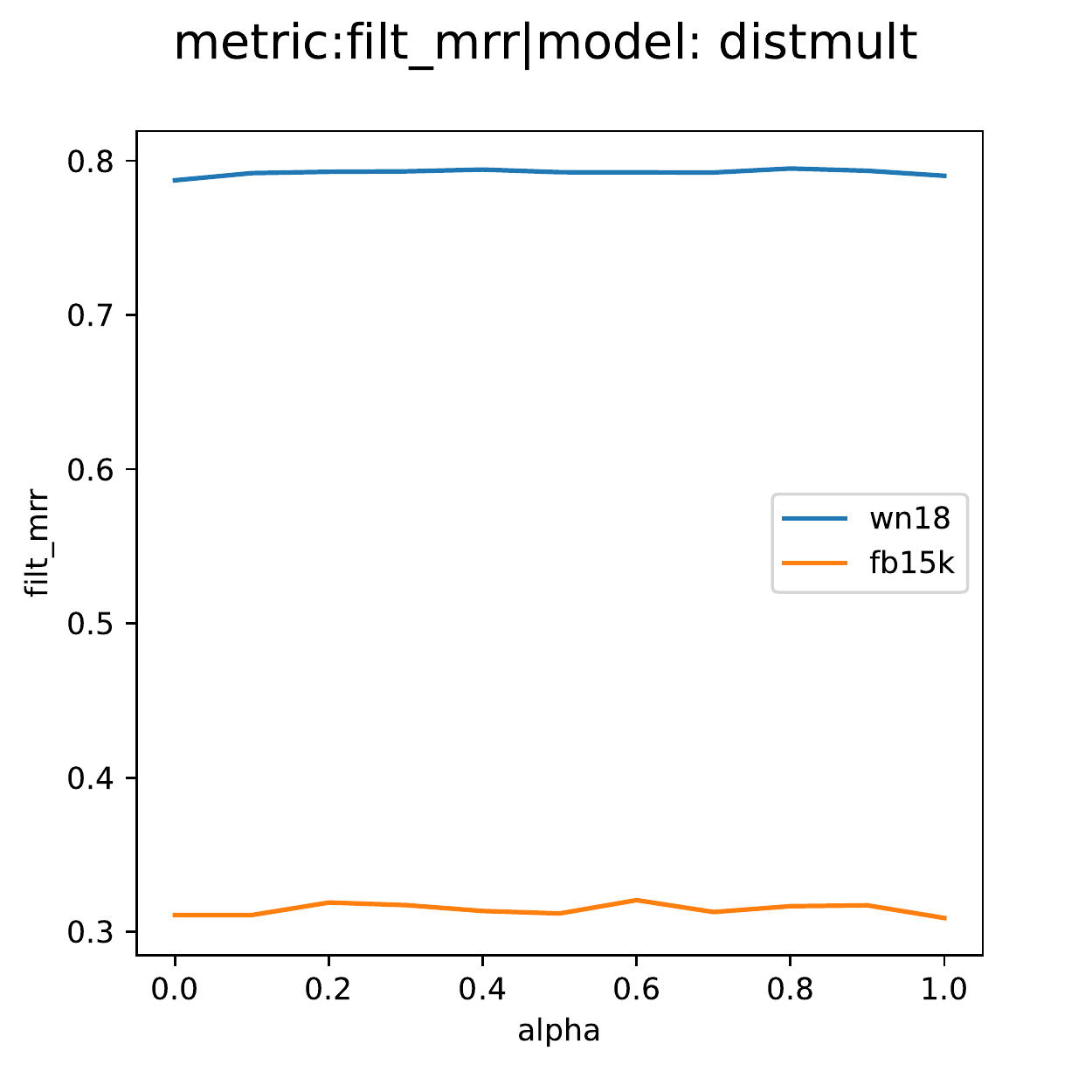}
					\caption{NSF-DistMult.}
					\label{fig:alpha_weighted_loss_distmult}
				\end{subfigure}
				\hfill
				\begin{subfigure}[b]{0.4\textwidth}
					\includegraphics[width=\textwidth]{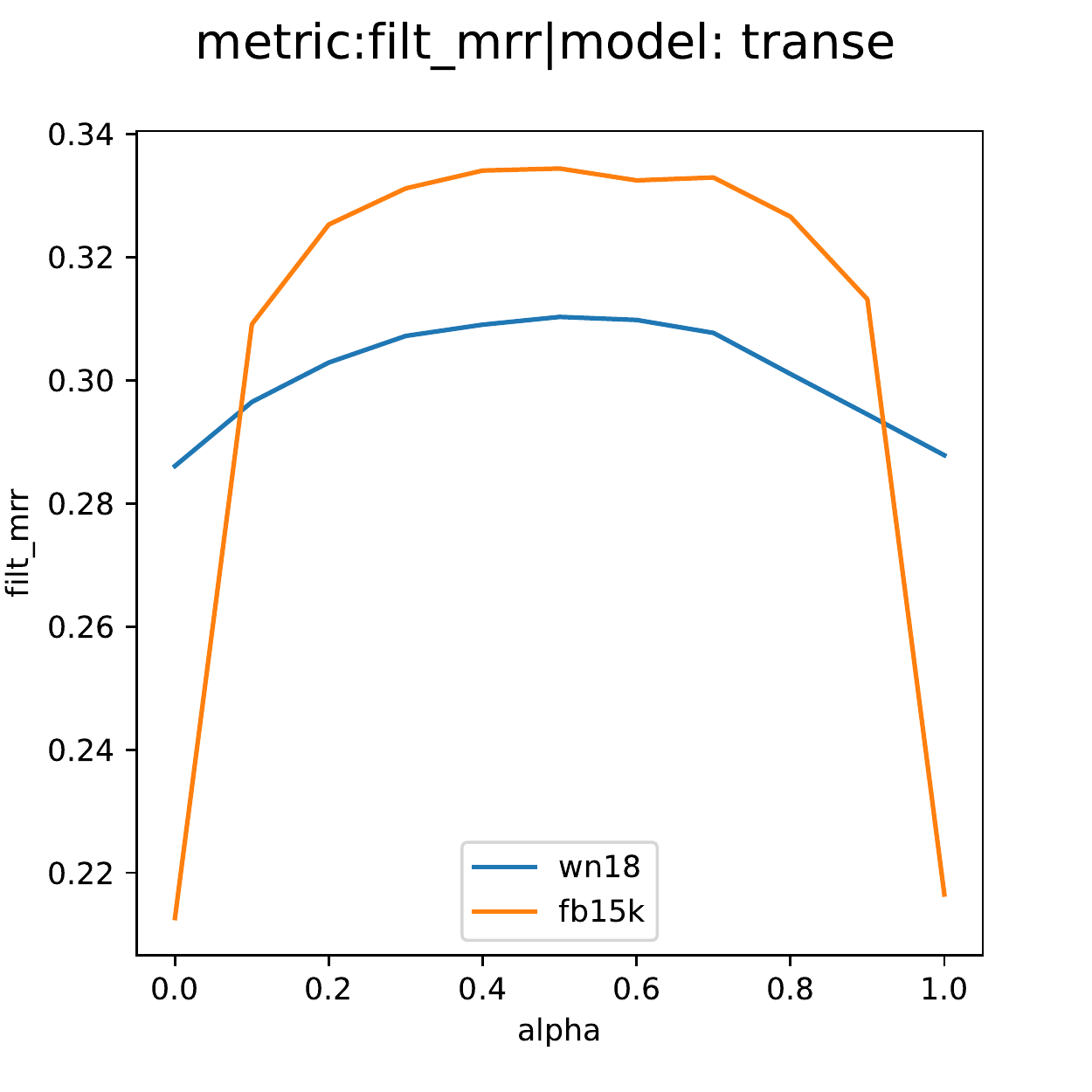}
					\caption{NSF-TransE.}
					\label{fig:alpha_weighted_loss_transe}
				\end{subfigure}
				\caption{The filtered $ MRR $ performance of NSF-DistMult (left) and NSF-TransE (right) as a function of varying $\alpha$ on the WN18 and FB15K datasets.}
				\label{fig:alpha_weighted_loss}
			\end{figure}

			The purpose of this ablation study is to explore the effects of using different weights ($ \alpha $ in equation \ref{eq:weightedLoss}) on the performance of the models on link prediction.
			\begin{equation}
				\label{eq:weightedLoss}
			\mathcal{L}(\{(h,r,t)\}_{i=1}^{b})=\alpha\mathcal{L}_{BT}(\mathbf{H}_{|},\mathbf{T})+(1-\alpha)\mathcal{L}_{BT}(\mathbf{H},\mathbf{T}_{|})
			\end{equation}
			We consider $ \alpha\in \{0,0.1,0.2,0.3,0.4,0.5,0.6,0.7,0.8,0.9,1\} $. In order to decrease the number of our experiments, we fixed the embedding dimension $ d $, learning rate $ lr $ and batch size $ b $ to $ d= 500$, $ b=2000 $ and $ lr=0.001 $ for NSF-DistMult, and to $ d= 500$, $ b=4000 $, $ lr=0.001 $ for NSF-TransE. We used $ L_{2} $ distance as a similarity measure. Figure \ref{fig:alpha_weighted_loss} shows performance as a function of $\alpha$. Figure \ref{fig:alpha_weighted_loss_distmult} shows that NSF-DistMult is not sensitive to $\alpha$, meaning that using only term is sufficient for this method. However, for NSF-TransE, figure \ref{fig:alpha_weighted_loss_transe} shows that the performance is impacted significantly $\alpha$. Optimum performance is obtained with $\alpha=0.5$, which corresponds to the expression in \ref{eq:tbt}. We can attribute this to the symmetric nature of DistMult score function and the asymmetric nature of TransE, which makes the inclusion of the second term essential.
		\subsection{HSIC}
			In this part, we are interested in studying the effect of replacing the BT loss in equation \ref{eq:tbt} with the HSIC loss (equation \ref{eq:hsic}) on the performance of the different models. The objective that we optimize in this ablation study is
			\begin{equation}
				\label{eq:thsic}
				\mathcal{L}(\{(h,r,t)\}_{i=1}^{b})=\mathcal{L}_{HSIC}(\mathbf{H}_{|},\mathbf{T})+\mathcal{L}_{HSIC}(\mathbf{H},\mathbf{T}_{|})
			\end{equation}
			We did not use ShuffleDBN in this experiment. We varied the dimension size, the learning rate, the batch size and the dissimilarity type (in the case of TransE) in a manner similar to what we described in section \ref{sec:expsetup}. Table \ref{tab:hsicperformance} shows the different results that we obtained. We notice that using HSIC instead of BT causes a slight decrease in performance for most metrics.
			
			\begin{table}[ht]
				\centering
				\scalebox{0.6}{
					\begin{tabular}{|c|c|c|c|c|c|c|c|c|c|c|c|c|c|c|c|c|}
						\hline
						&\multicolumn{8}{c}{FB15K}&\multicolumn{8}{|c|}{WN18}\\
						\cline{2-17}
						&\multicolumn{2}{c}{$ MR $}&\multicolumn{2}{|c}{$ MRR $}&\multicolumn{2}{|c}{$ H@1 $}&\multicolumn{2}{|c}{$ H@10 $}&\multicolumn{2}{|c}{$ MR $}&\multicolumn{2}{|c}{$ MRR $}&\multicolumn{2}{|c}{$ H@1 $}&\multicolumn{2}{|c|}{$ H@10 $}\\
						\cline{2-17}
						&raw& filt.&raw& filt.&raw& filt.&raw& filt.&raw& filt.&raw& filt.&raw& filt.&raw& filt.\\
						\hline
						NFS-TransE (w/ BT)&	\textbf{351}&	\textbf{256}&	\textbf{0.2117}&	\textbf{0.3366}&	\textbf{12.12}&	\textbf{23.70}&	\textbf{39.82}&	\textbf{50.89}&	\textbf{473}&	\textbf{460}&	\textbf{0.2388}&	\textbf{0.3149}&	\textbf{04.32}&	\textbf{07.21}&	68.08&	78.59\\
						\hline
						NFS-TransE (w/ HSIC)&	393& 	298& 	0.2015&	0.3250&	11.39&	22.73&	38.39&	49.23&	476&	463&	0.2387&	0.3119&	3.87&	6.17&	\textbf{68.63}&	\textbf{78.97}\\
						\hline
						\hline
						NFS-DistMult (w/ BT)&	\textbf{409}& 	\textbf{311}&	\textbf{0.1930}&	\textbf{0.3265}&	10.99&	\textbf{23.18}&	\textbf{36.75}&	\textbf{49.42}&	\textbf{641}&	\textbf{626}&	0.4944&	\textbf{0.7955}&	33.64&	\textbf{69.05}&	79.40&	\textbf{93.29}\\
						\hline
						NFS-DistMult (w/ HSIC)&\textbf{409}&	\textbf{311}&	0.1922&	0.3209&	\textbf{11.06}&	22.73&36.35&	48.69&	656&	642& 	\textbf{0.4964}&	0.7946&	\textbf{33.85}& 	\textbf{69.05}&	\textbf{79.56}&	93.14\\
						\hline
					\end{tabular}
				}
				\caption{Results on FB15K and WN18 for different losses.}
				\label{tab:hsicperformance}
			\end{table}
		\subsection{Effect of Loss Design from Score Function}
			In this ablation study, we aim to quantify the effect of having a loss function based on the score function (used for inference), as described in section \ref{sec:methodology}, on the performance of the model. More precisely, we trained a model,  Alt-NSF-TransE, with the loss function designed for DistMult described in equation \ref{eq:btdistmultLoss}, but with the TransE score function for inference(equation \ref{eq:transeSF}). We also trained  Alt-NSF-DistMult with the loss designed based on the TransE score function (equation \ref{eq:btTransELoss}), but inference is done using the DistMult score function (equation \ref{eq:distmultSF}). We varied the dimension size, the learning rate, the batch size and the dissimilarity type (in the case of TransE) in a manner similar to what we described in section \ref{sec:expsetup}. We do not employ the ShuffleDBN layer in these experiments. Table \ref{tab:altperformance} shows the results for different datasets. We notice that there is a significant drop in performance when using the loss design for an alternative model, which shows the strong connection that exists between the design of the loss function and the score function used during inference.
		\begin{table}[ht]
			\centering
			\scalebox{0.6}{
				\begin{tabular}{|c|c|c|c|c|c|c|c|c|c|c|c|c|c|c|c|c|}
					\hline
					&\multicolumn{8}{c}{FB15K}&\multicolumn{8}{|c|}{WN18}\\
					\cline{2-17}
					&\multicolumn{2}{c}{$ MR $}&\multicolumn{2}{|c}{$ MRR $}&\multicolumn{2}{|c}{$ H@1 $}&\multicolumn{2}{|c}{$ H@10 $}&\multicolumn{2}{|c}{$ MR $}&\multicolumn{2}{|c}{$ MRR $}&\multicolumn{2}{|c}{$ H@1 $}&\multicolumn{2}{|c|}{$ H@10 $}\\
					\cline{2-17}
					&raw& filt.&raw& filt.&raw& filt.&raw& filt.&raw& filt.&raw& filt.&raw& filt.&raw& filt.\\
					\hline
					NFS-TransE&	\textbf{351}&	\textbf{256}&	\textbf{0.2117}&	\textbf{0.3366}&	\textbf{12.12}&	\textbf{23.70}&	\textbf{39.82}&	\textbf{50.89}&	\textbf{473}&	\textbf{460}&	\textbf{0.2388}&	\textbf{0.3149}&	\textbf{04.32}&	\textbf{07.21}&	\textbf{68.08}&	\textbf{78.59}\\
					\hline
					NFS-TransE (Alt)&	8440&	8353&	0.014942 &		0.026535&	0.43&	0.58&	3.82&	5.57&	7143&	7134&	0.0476&	0.0543&	0.85&	1.02&	0.1318&	0.1433\\
					\hline
					\hline
					NFS- DistMult&	\textbf{409}& 	\textbf{311}&	\textbf{0.1930}&	\textbf{0.3265}&	\textbf{10.99}&	\textbf{23.18}&	\textbf{36.75}&	\textbf{49.42}&	\textbf{641}&	\textbf{626}&	\textbf{0.4944}&	\textbf{0.7955}&	\textbf{33.64}&	\textbf{69.05}&	\textbf{79.40}&	\textbf{93.29}\\
					\hline
					NFS- DistMult (Alt)&	7842&	7753&	0.005725&	0.009322&	0.20&	0.65&	1.10&	1.24&	9326&	9317&	0.0113&	0.0123&	0.31&	0.36&	2.46&	2.63\\
					\hline
				\end{tabular}
			}
			\caption{Results on FB15K and WN18 for different alternative losses.}
			\label{tab:altperformance}
		\end{table}
		
%
\vspace{-0.8cm}
\section{Conclusion}
	\label{sec:conclusion}
	Negative-sample-free link prediction is still at a nascent stage with many challenges remained unsolved. In this paper, we propose a novel approach (KG-NSF) to learn  embeddings for knowledge graph entities and relations for the purpose of link prediction without using negative examples. In particular, feature decorrelation is introduced to replace the need for negative examples, which leads to avoiding learning trivial solutions (i.e. representation collapse). Experimental results on multiple link prediction datasets have shown either a superior or comparable performance of our approach. This work demonstrates the possibility of learning knowledge graph embeddings without negative sampling and without explicit constraints on the score functions. One potential limitation of our approach is that it is memory intensive, since it requires multiplications between large matrices.

{
\small
\bibliographystyle{unsrt}
\medskip
\bibliography{ms.bib}

}


\appendix
\section{Implementation Details}
\begin{itemize}
	\item Main code: Our implementation is built on top of the code of the TorchKGE package available at \url{https://torchkge.readthedocs.io/en/latest/}. We customized our training loop so that it can stop after it reaches the optimal validation filtered $ MRR $ metric, with a patience of 5 epochs. We used the the same link prediction evaluation module available\footnote{\url{https://torchkge.readthedocs.io/en/latest/reference/evaluation.html}} in TorchKGE with some minor changes. We adapted the TransE and DistMult models that exist in TorchKGE so that they don't have any negative sampling and so that the "forward" function contains the NSF loss.
	\item ShuffleDBN layer: We used the same implementation available at \url{https://github.com/PatrickHua/FeatureDecorrelationSSL/blob/main/models/utils/pseudo_code_for_paper.py}.
	\item Datasets:
	\begin{itemize}
		\item FB15K: We used the dataset available via the TorchKGE package \url{https://torchkge.readthedocs.io/en/latest/reference/utils.html}. We used the default split.
		\item WN18: We used the dataset available via the TorchKGE package. We used the default split. 
		\item FB15k-237AM: We used the dataset available via the Github repository of \cite{hajimoradlou2022stay} at \url{https://github.com/BorealisAI/StayPositive/tree/master/dataset/FB15k-237AM}. We used the default split.
		\item WN18AM: We used the dataset available via the Github repository of \cite{hajimoradlou2022stay} at \url{https://github.com/BorealisAI/StayPositive/tree/master/dataset/WN18AM}. We used the default split.
	\end{itemize}
\end{itemize}

\section{NSF-TransE Vs. TransE}
\label{sec:nfstransEDER}
In what follows $ \text{tr}(\mathbf{A}) $ refers to the trace of matrix $\mathbf{A}$. We used some standard properties of the matrix trace in this section such as
\begin{equation*}\label{key}
\operatorname {tr} \left(\mathbf {A} ^{\mathsf {T}}\mathbf {B} \right)=\operatorname {tr} \left(\mathbf {A} \mathbf {B} ^{\mathsf {T}}\right)=\operatorname {tr} \left(\mathbf {B} ^{\mathsf {T}}\mathbf {A} \right)=\operatorname {tr} \left(\mathbf {B} \mathbf {A} ^{\mathsf {T}}\right)=\sum _{i=1}^{m}\sum _{j=1}^{n}a_{ij}b_{ij}\
\end{equation*}
This means that $ \operatorname{tr}\left(\mathbf{A}^{2}\right)=\sum _{i=1}^{m}\sum _{j=1}^{n}a_{ij}^{2}$.
In what follows we relate TransE objective and our objective via non-standardized empirical cross-correlation matrix $\widetilde{\mathcal{C}}=\text{cor}\left(\mathbf{X},\mathbf{Y}\right)=\mathbf{X}^{\top}\mathbf{Y}$, where $ \widetilde{\mathcal{C}}_{ij} = \sum_{b}x_{b,i}y_{b,j} $, as opposed to the standardized empirical cross-correlation whose elements are expressed as follows: $ \mathcal{C}_{ij}=\frac{\sum_{b}x_{b,i}y_{b,j}}{\sqrt{\sum_{b}(x_{b,i})^{2}}\sqrt{\sum_{b}(y_{b,j})^{2}}} $. Even though the two expressions are different but from an optimization prospective they share similar behavior. The loss function of TransE for positive triples is (in the case of L2 norm):
\begin{equation}\label{key}
\mathcal{L}^{+}=\sum_{i}^{b}\|\mathbf{h}_{i}+\mathbf{r}_{i}-\mathbf{t}_{i}\|_{2}=\sum_{i}^{b}(\mathbf{h}_{i}+\mathbf{r}_{i}-\mathbf{t}_{i})^{\top}(\mathbf{h}_{i}+\mathbf{r}_{i}-\mathbf{t}_{i})\\
\end{equation}

This loss can be reformulated in the following three forms:
\begin{equation}\label{key}
\mathcal{L}^{+}_{1}=\sum_{i}^{b}(\mathbf{h}_{i}+\mathbf{r}_{i})^{\top}(\mathbf{h}_{i}+\mathbf{r}_{i})-2(\mathbf{h}_{i}+\mathbf{r}_{i})^{\top}\mathbf{t}_{i}+\mathbf{t}_{i}^{\top}\mathbf{t}_{i}
\end{equation}
\begin{equation}\label{key}
\mathcal{L}_{2}^{+}=\sum_{i}^{b}\mathbf{h}_{i}^{\top}\mathbf{h}_{i}-2\mathbf{h}_{i}^{\top}\left(\mathbf{t}_{i}-\mathbf{r}_{i}\right)+\left(\mathbf{t}_{i}-\mathbf{r}_{i}\right)^{\top}\left(\mathbf{t}_{i}-\mathbf{r}_{i}\right)
\end{equation}
\begin{equation}\label{key}
\mathcal{L}_{3}^{+}=\sum_{i}^{b}(\mathbf{h}_{i}-\mathbf{t}_{i})^{\top}(\mathbf{h}_{i}-\mathbf{t}_{i})+2(\mathbf{h}_{i}-\mathbf{t}_{i})^{\top}\mathbf{r}_{i}+\mathbf{r}_{i}^{\top}\mathbf{r}_{i}
\end{equation}
It is easy to show that $\mathcal{L}^{+}=\mathcal{L}^{+}_{1}=\mathcal{L}^{+}_{2}=\mathcal{L}^{+}_{3}$.

This can be presented in a matrix form as follows: let $ \mathbf{H}=[\mathbf{h}_{i}]_{i=1}^{b}\in\mathbb{R}^{b\times d} $,  $ \mathbf{T}=[\mathbf{t}_{i}]_{i=1}^{b}\in\mathbb{R}^{b\times d} $, $ \mathbf{R}=[\mathbf{r}_{i}]_{i=1}^{b}\in\mathbb{R}^{b\times d} $, $\mathbf{H}_{|}=\mathbf{H}+\mathbf{R}\in\mathbb{R}^{b\times d}$, $\mathbf{T}_{|}=\mathbf{T}-\mathbf{R}\in\mathbb{R}^{b\times d}$ and $\mathbf{R}_{|}=\mathbf{H}-\mathbf{T}$.
\begin{equation}\label{key}
\mathcal{L}_{1}^{+}=\text{tr}\left(\mathbf{H}_{|}\mathbf{H}_{|}^{\top}\right)-2\text{tr}\left(\mathbf{H}_{|}\mathbf{T}^{\top}\right)+\text{tr}\left(\mathbf{T}\mathbf{T}^{\top}\right)
\end{equation}
\begin{equation}\label{key}
\mathcal{L}_{2}^{+}=\text{tr}\left(\mathbf{H}\mathbf{H}^{\top}\right)-2\text{tr}\left(\mathbf{H}\mathbf{T}_{|}^{\top}\right)+\text{tr}\left(\mathbf{T}_{|}\mathbf{T}_{|}^{\top}\right)
\end{equation}
\begin{equation}\label{key}
\mathcal{L}_{3}^{+}=\text{tr}\left(\mathbf{R}_{|}\mathbf{R}_{|}^{\top}\right)-2\text{tr}\left(\mathbf{R}\mathbf{R}_{|}^{\top}\right)+\text{tr}\left(\mathbf{R}\mathbf{R}^{\top}\right)
\end{equation}

Additionally we have:
\begin{equation}\label{key}
\begin{split}
\mathbf{H}_{|}\mathbf{H}_{|}^{\top}&=\left(\mathbf{H}+\mathbf{R}\right)\left(\mathbf{H}+\mathbf{R}\right)^{\top}\\
&=\left(\mathbf{H}+\mathbf{R}\right)\left(\mathbf{H}^{\top}+\mathbf{R}^{\top}\right)\\
&=\mathbf{H}\mathbf{H}^{\top}+\mathbf{H}\mathbf{R}^{\top}+\mathbf{R}\mathbf{H}^{\top}+\mathbf{R}\mathbf{R}^{\top}\\
\end{split}
\end{equation}
and
\begin{equation}\label{key}
\begin{split}
\mathbf{T}_{|}\mathbf{T}_{|}^{\top}&=\left(\mathbf{T}-\mathbf{R}\right)\left(\mathbf{T}-\mathbf{R}\right)^{\top}\\
&=\left(\mathbf{T}-\mathbf{R}\right)\left(\mathbf{T}^{\top}-\mathbf{R}^{\top}\right)\\
&=\mathbf{R}\mathbf{R}^{\top}-\mathbf{R}\mathbf{T}^{\top}-\mathbf{T}\mathbf{R}^{\top}+\mathbf{T}\mathbf{T}^{\top}\\
\end{split}
\end{equation}
and
\begin{equation}\label{key}
\begin{split}
\mathbf{R}_{|}\mathbf{R}_{|}^{\top}&=\left(\mathbf{H}-\mathbf{T}\right)\left(\mathbf{H}-\mathbf{T}\right)^{\top}\\
&=\left(\mathbf{H}-\mathbf{T}\right)\left(\mathbf{H}^{\top}-\mathbf{T}^{\top}\right)\\
&=\mathbf{H}\mathbf{H}^{\top}-\mathbf{H}\mathbf{T}^{\top}-\mathbf{T}\mathbf{H}^{\top}+\mathbf{T}\mathbf{T}^{\top}\\
\end{split}
\end{equation}
Consequently, $ \mathcal{L}^{+}_{1} $ can be expressed as follows,
\begin{equation}\label{key}
\begin{split}
\mathcal{L}^{+}_{1}&=\text{tr}\left(\mathbf{H}\mathbf{H}^{\top}+\mathbf{H}\mathbf{R}^{\top}+\mathbf{R}\mathbf{H}^{\top}+\mathbf{R}\mathbf{R}^{\top}\right)-2\text{tr}\left(\mathbf{H}_{|}\mathbf{T}^{\top}\right)+\text{tr}\left(\mathbf{T}\mathbf{T}^{\top}\right)\\
&=\text{tr}\left(\mathbf{H}\mathbf{H}^{\top}\right)+\text{tr}\left(\mathbf{H}\mathbf{R}^{\top}\right)+\text{tr}\left(\mathbf{R}\mathbf{H}^{\top}\right)+\text{tr}\left(\mathbf{R}\mathbf{R}^{\top}\right)-2\text{tr}\left(\mathbf{H}_{|}\mathbf{T}^{\top}\right)+\text{tr}\left(\mathbf{T}\mathbf{T}^{\top}\right)\\
&=\text{tr}\left(\mathbf{H}\mathbf{H}^{\top}\right)+\text{tr}\left(\mathbf{R}\mathbf{R}^{\top}\right)+\text{tr}\left(\mathbf{T}\mathbf{T}^{\top}\right)+2\text{tr}\left(\mathbf{R}\mathbf{H}^{\top}\right)-2\text{tr}\left(\mathbf{H}_{|}\mathbf{T}^{\top}\right)\\
&=\|\mathbf{H}\|_{2}^{2}+\|\mathbf{R}\|_{2}^{2}+\|\mathbf{T}\|_{2}^{2}+2\text{tr}\left(\mathbf{R}\mathbf{H}^{\top}\right)-2\text{tr}\left(\mathbf{H}_{|}\mathbf{T}^{\top}\right)\\
\end{split}
\end{equation}
$ \mathcal{L}^{+}_{2} $ can be expressed as follows:
\begin{equation}\label{key}
\begin{split}
\mathcal{L}^{+}_{2}&=\text{tr}\left(\mathbf{H}\mathbf{H}^{\top}\right)-2\text{tr}\left(\mathbf{H}\mathbf{T}_{|}^{\top}\right)+\text{tr}\left(\mathbf{T}_{|}\mathbf{T}_{|}^{\top}\right)\\
&=\text{tr}\left(\mathbf{H}\mathbf{H}^{\top}\right)-2\text{tr}\left(\mathbf{H}\mathbf{T}_{|}^{\top}\right)+\text{tr}\left(\mathbf{R}\mathbf{R}^{\top}-\mathbf{R}\mathbf{T}^{\top}-\mathbf{T}\mathbf{R}^{\top}+\mathbf{T}\mathbf{T}^{\top}\right)\\
&=\text{tr}\left(\mathbf{H}\mathbf{H}^{\top}\right)-2\text{tr}\left(\mathbf{H}\mathbf{T}_{|}^{\top}\right)+
\text{tr}\left(\mathbf{R}\mathbf{R}^{\top}\right)-\text{tr}\left(\mathbf{R}\mathbf{T}^{\top}\right)-\text{tr}\left(\mathbf{T}\mathbf{R}^{\top}\right)+\text{tr}\left(\mathbf{T}\mathbf{T}^{\top}\right)\\
&=\text{tr}\left(\mathbf{H}\mathbf{H}^{\top}\right)+\text{tr}\left(\mathbf{R}\mathbf{R}^{\top}\right)+\text{tr}\left(\mathbf{T}\mathbf{T}^{\top}\right)-2\text{tr}\left(\mathbf{R}\mathbf{T}^{\top}\right)-2\text{tr}\left(\mathbf{H}\mathbf{T}_{|}^{\top}\right)
\end{split}
\end{equation}
$ \mathcal{L}^{+}_{3} $ can be expressed as follows:
\begin{equation}\label{key}
\begin{split}
\mathcal{L}^{+}_{3}&=\text{tr}\left(\mathbf{R}_{|}\mathbf{R}_{|}^{\top}\right)+2\text{tr}\left(\mathbf{R}\mathbf{R}_{|}^{\top}\right)+\text{tr}\left(\mathbf{R}\mathbf{R}^{\top}\right)\\
&=\text{tr}\left(\mathbf{H}\mathbf{H}^{\top}-\mathbf{H}\mathbf{T}^{\top}-\mathbf{T}\mathbf{H}^{\top}+\mathbf{T}\mathbf{T}^{\top}\right)+2\text{tr}\left(\mathbf{R}\mathbf{R}_{|}^{\top}\right)+\text{tr}\left(\mathbf{R}\mathbf{R}^{\top}\right)\\
&=\text{tr}\left(\mathbf{H}\mathbf{H}^{\top}\right)-\text{tr}\left(\mathbf{H}\mathbf{T}^{\top}\right)-\text{tr}\left(\mathbf{T}\mathbf{H}^{\top}\right)+\text{tr}\left(\mathbf{T}\mathbf{T}^{\top}\right)+2\text{tr}\left(\mathbf{R}\mathbf{R}_{|}^{\top}\right)+\text{tr}\left(\mathbf{R}\mathbf{R}^{\top}\right)\\
&=\text{tr}\left(\mathbf{H}\mathbf{H}^{\top}\right)+\text{tr}\left(\mathbf{R}\mathbf{R}^{\top}\right)+\text{tr}\left(\mathbf{T}\mathbf{T}^{\top}\right)-2\text{tr}\left(\mathbf{T}\mathbf{H}^{\top}\right)+2\text{tr}\left(\mathbf{R}\mathbf{R}_{|}^{\top}\right)\\
&=\|\mathbf{H}\|_{2}^{2}+\|\mathbf{R}\|_{2}^{2}+\|\mathbf{T}\|_{2}^{2}-2\text{tr}\left(\mathbf{T}\mathbf{H}^{\top}\right)+2\text{tr}\left(\mathbf{R}\left(\mathbf{H}-\mathbf{T}\right)^{\top}\right)
\end{split}
\end{equation}
If we combine the three expressions of $ \mathcal{L}^{+}_{\text{TransE}} $ we have:
\begin{equation}
\label{eq:transeBTDecomp}
\begin{split}
&\mathcal{L}^{+}_{\text{TransE}}=\frac{1}{3}\left(\mathcal{L}^{+}_{1}+\mathcal{L}^{+}_{2}+\mathcal{L}^{+}_{3}\right)\\
&=\nu+\frac{2}{3}\text{tr}\left(\mathbf{R}\mathbf{H}^{\top}\right)-\frac{2}{3}\text{tr}\left(\mathbf{R}\mathbf{T}^{\top}\right)+\frac{2}{3}\text{tr}\left(\mathbf{R}\left(\mathbf{H}-\mathbf{T}\right)^{\top}\right)-\frac{2}{3}\text{tr}\left(\mathbf{H}_{|}\mathbf{T}^{\top}\right)-\frac{2}{3}\text{tr}\left(\mathbf{H}\mathbf{T}_{|}^{\top}\right)-\frac{2}{3}\text{tr}\left(\mathbf{T}\mathbf{H}^{\top}\right)\\
&=\nu+\frac{4}{3}\text{tr}\left(\mathbf{R}\left(\mathbf{H}-\mathbf{T}\right)^{\top}\right)-\frac{2}{3}\text{tr}\left(\mathbf{H}_{|}\mathbf{T}^{\top}\right)-\frac{2}{3}\text{tr}\left(\mathbf{H}\mathbf{T}_{|}^{\top}\right)-\frac{2}{3}\text{tr}\left(\mathbf{T}\mathbf{H}^{\top}\right)\\
&=\nu+\frac{4}{3}\text{tr}\left(\mathbf{R}^{\top}\left(\mathbf{H}-\mathbf{T}\right)\right)-\frac{2}{3}\text{tr}\left(\mathbf{H}_{|}^{\top}\mathbf{T}\right)-\frac{2}{3}\text{tr}\left(\mathbf{H}^{\top}\mathbf{T}_{|}\right)-\frac{2}{3}\text{tr}\left(\mathbf{T}^{\top}\mathbf{H}\right)\\
&=\nu+\frac{4}{3}\text{tr}\left[\text{cor}(\mathbf{R},\left(\mathbf{H}-\mathbf{T}\right))\right]-\frac{2}{3}\text{tr}\left[\text{cor}(\mathbf{H}_{|},\mathbf{T})\right]-\frac{2}{3}\text{tr}\left[\text{cor}(\mathbf{H},\mathbf{T}_{|})\right]-\frac{2}{3}\text{tr}\left[\text{cor}(\mathbf{T},\mathbf{H})\right]
\end{split}
\end{equation}
With $\nu=\|\mathbf{H}\|_{2}^{2}+\|\mathbf{R}\|_{2}^{2}+\|\mathbf{T}\|_{2}^{2}$.
By analogy, the expression of the loss on the negative examples (for an arbitrary number of negative examples per positive example) can be expressed as follows (note that it should be maximized),
\begin{equation}\label{key}
\mathcal{L}^{-}_{\text{TransE}}=\frac{2}{3}\text{tr}\left[\text{cor}(\mathbf{H}_{|}',\mathbf{T}')\right]+\frac{2}{3}\text{tr}\left[\text{cor}(\mathbf{H}',\mathbf{T}_{|}')\right]+\frac{2}{3}\text{tr}\left[\text{cor}(\mathbf{T}',\mathbf{H}')\right]-\nu-\frac{4}{3}\text{tr}\left[\text{cor}(\mathbf{R}',\left(\mathbf{H}'-\mathbf{T}'\right))\right]
\end{equation}

The objective is to minimize the previous score function on positive examples, which is equivalent to maximizing the on-diagonal elements of $ \text{cor}(\mathbf{H}_{|},\mathbf{T}) $ and $ \text{cor}(\mathbf{H},\mathbf{T}_{|}) $, which is similar to what our loss does. In fact, $\mathcal{L}_{BT}(\mathbf{H}_{|},\mathbf{T})$ maximizes the on-diagonal elements of $ \text{cor}(\mathbf{H}_{|},\mathbf{T}) $ and minimizes its off-diagonal terms.  $\mathcal{L}_{BT}(\mathbf{H},\mathbf{T}_{|})$ maximizes the on-diagonal elements of $ \text{cor}(\mathbf{H},\mathbf{T}_{|}) $ and minimizes its off-diagonal terms. We hypothesize that our approach avoids representation collapse by minimizing the off-diagonal terms. Concerning the second and 5th terms in equation \ref{eq:transeBTDecomp}, we implemented the following objective, which satisfies them:
\begin{equation}\label{key}
\mathcal{L}=\mathcal{L}(\mathbf{H}_{|},\mathbf{T})+\mathcal{L}(\mathbf{H},\mathbf{T}_{|})+\mathcal{L}(\mathbf{H},\mathbf{T})-\mathcal{L}(\mathbf{R},\mathbf{H}-\mathbf{T})
\end{equation}
This objective was also consistently minimized and the additional constraints didn't cause any significant collapse, but the performance of our model deteriorated slightly. This deterioration in the performance after minimizing the additional losses can be attributed to the dimensions that are  decorrelated in these terms, which shouldn't be decorrelated. In fact, $\mathbf{H}$ and $\mathbf{T}$ may have the same elements (since the tail and the head can be interchangeable), consequently decorrelating them is counter intuitive. This behavior is also noticed in the case of negative sample free contrastive learning \cite{hua2021feature}. \cite{hua2021feature} observed that increasing feature decorrelation can improve the performance of models. This shows empirically the sufficiency of $ \mathcal{L}(\mathbf{H}_{|},\mathbf{T})+\mathcal{L}(\mathbf{H},\mathbf{T}_{|}) $.

\section{NSF-Distmult Vs. Distmult}
\label{sec:nfsdistmutDER}
Distmult score function is:
\begin{equation}
\label{eq:}
f(h,r,t)=\sum_{i}\left[\mathbf{h}\odot \mathbf{r} \odot \mathbf{t}\right]_{i}
\end{equation}
And its positive score function can be expressed as follows:
\begin{equation}\label{key}
\mathcal{L}^{+}_{\text{Distmult}}=\sum_{j=1}^{b}f(h_{j},r_{j},t_{j})
\end{equation}
Let $\mathbf{T}_{|}=\mathbf{T}\odot\mathbf{R}$, $\mathbf{H}_{|}=\mathbf{H}\odot\mathbf{R}$ and $\mathbf{R}_{|}=\mathbf{H}\odot\mathbf{T}$. $\odot$ is the Hadamard product. This can be expressed in matrix form as follows:
\begin{equation}
\label{eq:}
\begin{split}
\mathcal{L}^{+}_{\text{Distmult}}&=\sum_{i=1}^{b}\sum_{j=1}^{d}\left[\mathbf{H}\odot\mathbf{R}\odot\mathbf{T}\right]_{ij}\\
&=\sum_{i=1}^{b}\sum_{j=1}^{d}\mathbf{h}_{ij}\cdot \mathbf{r}_{ij}\cdot \mathbf{t}_{ij} \\
\end{split}
\end{equation}
This from the previous expression we can deduce that the following expressions are equal:
\begin{equation}
\label{eq:}
\begin{split}
\mathcal{L}_{1}^{+}&=\sum_{i=1}^{b}\sum_{j=1}^{d}\left(\mathbf{h}_{ij}\cdot \mathbf{r}_{ij}\right)\cdot \mathbf{t}_{ij} \\
&=\text{tr}\left(\mathbf{H}_{|}\mathbf{R}^{\top}\right)
\end{split}
\end{equation}
\begin{equation}
\label{eq:}
\begin{split}
\mathcal{L}_{2}^{+}&=\sum_{i=1}^{b}\sum_{j=1}^{d}\mathbf{h}_{ij}\cdot\left( \mathbf{r}_{ij}\cdot \mathbf{t}_{ij}\right) \\
&=\text{tr}\left(\mathbf{T}_{|}\mathbf{R}^{\top}\right)
\end{split}
\end{equation}
\begin{equation}
\label{eq:}
\begin{split}
\mathcal{L}_{3}^{+}&=\sum_{i=1}^{b}\sum_{j=1}^{d}\left(\mathbf{h}_{ij}\cdot \mathbf{t}_{ij}\right)\cdot \mathbf{r}_{ij} \\
&=\text{tr}\left(\mathbf{R}_{|}\mathbf{R}^{\top}\right)
\end{split}
\end{equation}
We can rewrite $\mathcal{L}^{+}$ as follows:
\begin{equation}\label{key}
\begin{split}
\mathcal{L}^{+}_{\text{Distmult}}&=\frac{1}{3}\left(\mathcal{L}^{+}_{1}+\mathcal{L}^{+}_{2}+\mathcal{L}^{+}_{3}\right)\\
&=\frac{1}{3}\text{tr}\left(\mathbf{H}_{|}\mathbf{R}^{\top}\right)+\frac{1}{3}\text{tr}\left(\mathbf{T}_{|}\mathbf{R}^{\top}\right)+\frac{1}{3}\text{tr}\left(\mathbf{R}_{|}\mathbf{R}^{\top}\right)
\end{split}
\end{equation}
We suppose that $\mathbf{H}_{|}$, $\mathbf{H}$, $\mathbf{T}$ and $\mathbf{T}_{|}$ are batch normalized i.e. their matrix product equals the empirical cross-correlation matrix. Then we will have,
\begin{equation}\label{key}
\mathcal{L}^{+}_{\text{Distmult}}=\frac{1}{3}\text{tr}\left[\text{cor}\left(\mathbf{H}_{|},\mathbf{R}\right)\right]+\frac{1}{3}\text{tr}\left[\text{cor}\left(\mathbf{T}_{|},\mathbf{R}\right)\right]+\frac{1}{3}\text{tr}\left[\text{cor}\left(\mathbf{R}_{|},\mathbf{R}\right)\right]
\end{equation}
We observe that minimizing the first two terms is equivalent to minimizing the on-diagonal terms of the objective that we used to train NSF-Distmult. The last term is not included in our loss since it has been shown to deteriorate performance.
\section{Optimal Hyper-parameters for Different Models and Datasets.}
Table \ref{tab:optConf} shows the optimal parameters obtained in various experiments. The optimality criteria of these models is the filtered $ MRR $ metric.
\begin{table}[ht]
	\centering
	\scalebox{0.7}{
		\begin{tabular}{|c|c|c|c|c|c|}
			\hline
			Dataset&Model&\multicolumn{4}{c|}{Optimal parameters}\\
			\cline{3-6}
			&&lr & d&b&dissimilarity type\\
			\hline
			\multirow{4}{*}{FB15K}&NSF-TransE (w/o SDBN)	&0.001&	500&	4000&	L2\\
			\cline{2-6}
			&NSF-DistMult (w/o SDBN)&	0.001&	500&	4000&	\_\\
			\cline{2-6}
			&NSF-TransE (w/ SDBN)&	0.0005&	400&	4000&	L2\\
			\cline{2-6}
			&NSF-DistMult (w/ SDBN)&	0.001&	500&	4000&	\_\\
			\hline
			\multirow{4}{*}{WN18}&NSF-TransE (w/o SDBN)&	0.001&	500&	1000&	L1\\
			
			\cline{2-6}
			&NSF-DistMult (w/o SDBN)&	0.0001&	500&	500&	\_\\
			\cline{2-6}
			&NSF-TransE (w/ SDBN)&	0.0005&	50& 	4000&	L1\\
			\cline{2-6}
			&NSF-DistMult (w/ SDBN)&	0.0001& 		500 &	4000 &	\_\\
			\hline
			\multirow{4}{*}{FB15k-237AM}&NSF- DistMult (w/o SDBN)&	0.0005&	400&	200&	\_\\
			\cline{2-6}
			&NSF- DistMult (w/ SDBN)&	0.001&	500&	200&	L2\\
			\cline{2-6}
			&NSF- TransE (w/o SDBN)&0.001&	500&	1000&	L2\\
			\cline{2-6}
			&NSF- TransE (w/ SDBN)&	0.0001&	500&	200&	L2\\
			\hline
			\multirow{4}{*}{WN18AM}&NSF- DistMult (w/o SDBN)&	0.0001&	400&	4000&	\_\\
			\cline{2-6}
			&NSF- DistMult (w/ SDBN)&	0.0001&	500&	4000&	\_\\
			\cline{2-6}
			&NSF- TransE (w/o SDBN)&0.0005&	100&	200&	L1\\
			\cline{2-6}
			&NSF- TransE (w/ SDBN)&	0.0010&	50&	1000&	L2\\
			\hline
	\end{tabular}}
	\caption{Optimal parameter configurations.}
	\label{tab:optConf}
\end{table}

\section{Hyperparameter Analysis}
\begin{figure}[ht]
	\includegraphics[width=\textwidth]{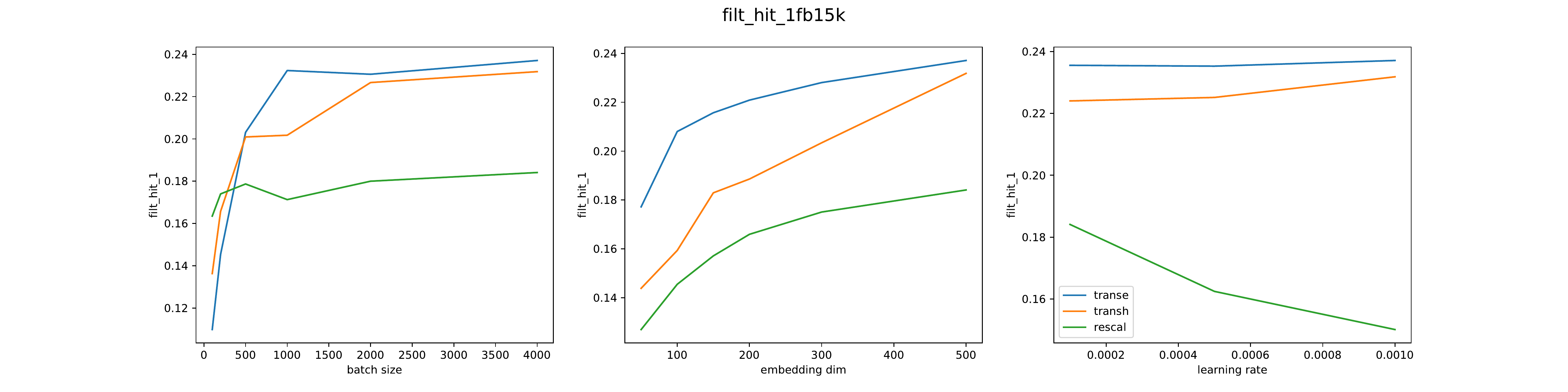}
	\caption{This figure shows the $ H@1 $ performance of the different models trained using the FB15K dataset.}
	\label{fig:filt_hit_1_fb15k}
\end{figure}

\begin{figure}[ht]
	\includegraphics[width=\textwidth]{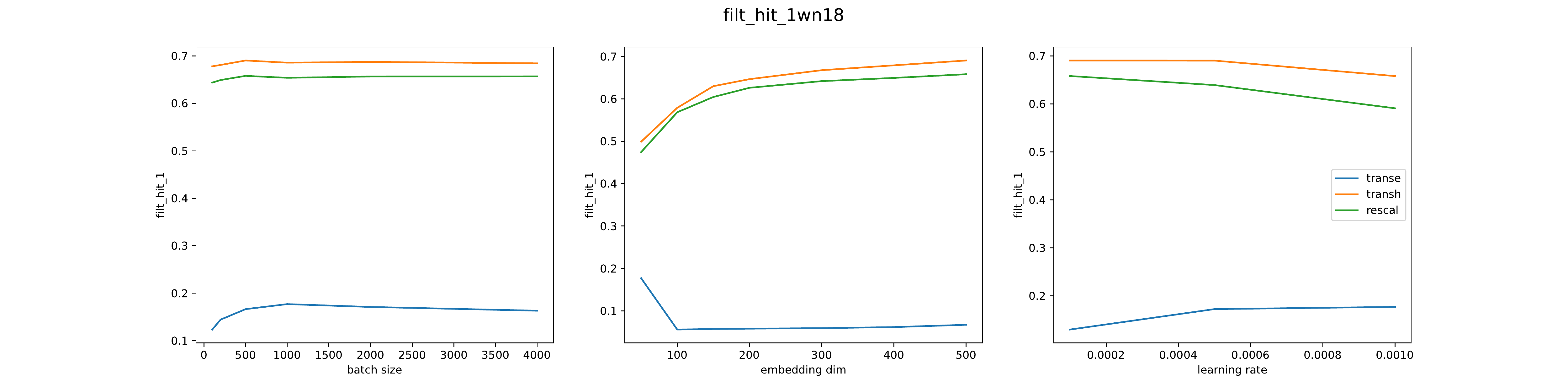}
	\caption{This figure shows the $ H@1 $ performance of the different models trained using the wn18 dataset.}
	\label{fig:filt_hit_1_wn18}
\end{figure}

In this part we try to see if there is a relationship between the different hyper-parameters that we used for our experiments and the performance of the models. Figures \ref{fig:filt_hit_1_fb15k} and \ref{fig:filt_hit_1_wn18} show the $ H@1 $ performance of TransE, DistMult and Rescal models. This is done as a function of either the batch size, the embedding dimension or the learning rate. In order to study each hyperparameter independent of the other hyperparameters, we fix the other hyperparameters using their respective values that had the best performance on the filtered $ MRR $ metric. We observe that increasing the batch size had a positive impact on the performance of the models across the different datasets. On the other hand the effect of the embedding size varies across datasets. In fact, for the FB15K dataset increasing the embedding dimension consistently improved the performance of the models, but in the case of the WN18 dataset increasing the embedding dimension improved DistMult and Rescal while deteriorating the performance of TransE. Similarly, the effect of the learning rate was inconsistent across datasets and models. In the case of the FB15K dataset increasing the learning rate improved the performance of TransE and DistMult slightly while deteriorating the performance of Rescal significantly. In the case of models trained using WN18, increasing the learning rate deteriorated the performance of Rescal and DistMult, while improving the performance of TransE.

\section{Convergence Analysis}
\begin{figure}[ht]
	\includegraphics[width=\textwidth]{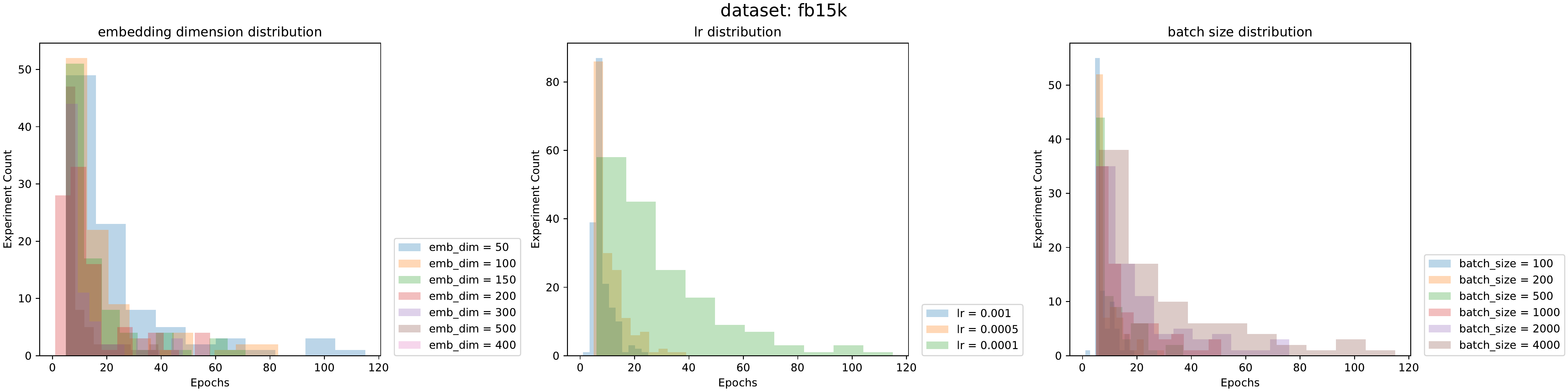}
	\caption{Distribution of the number of epochs needed for convergence on the set of conducted experiments on all the models (TransE, DistMult and Rescal).}
	\label{fig:fb15k}
\end{figure}

As was stated previously, our method increases the speed of convergence significantly relative to methods that use negative sampling. Figure \ref{fig:fb15k} shows the distribution of the number of epochs needed for convergence of different models trained of FB15K dataset. Each sub-figure is composed out of superposed histograms, and each histogram depends on a hyperparameter value. Taking multiple histograms depending on the hyperparameters was employed to understand the impact of the different hyperparameters on the overall convergence of the different models. First, we observe that the vast majority of models in all the experiments have converged in the first 40 epochs as apposed to models trained using negative sampling which need an order of a 100 epoch to converge \cite{rehman2022knowledge, mohamed2021training} (i.e. the number of epochs is $ 100\times k, k\in \mathbb{N} $). 

\end{document}